\pdfoutput=1

\documentclass{article}
\usepackage{ijcai25}

\usepackage{times}
\usepackage{soul}
\usepackage{url}
\usepackage[hidelinks]{hyperref}
\usepackage[utf8]{inputenc}
\usepackage[small]{caption}
\usepackage{graphicx}
\usepackage{amsmath}
\usepackage{amsthm}
\usepackage{booktabs}
\usepackage{algorithm}
\usepackage{algorithmic}
\usepackage[switch]{lineno}
\usepackage{amssymb}
\graphicspath{ {figure} }
\usepackage{multirow}
\usepackage[table,xcdraw]{xcolor}
\usepackage[normalem]{ulem}
\useunder{\uline}{\ul}{}
\usepackage{colortbl}
\usepackage{float}
\usepackage{subfigure}
\usepackage{color, xcolor}
\usepackage{soul}
\usepackage{xcolor}

\definecolor{darkblue}{HTML}{7195C5}
\definecolor{lightblue}{HTML}{D0D8EB}
\definecolor{darkpurple}{HTML}{414986}
\definecolor{lightpurple}{HTML}{C2BDDB}
\definecolor{darkorange}{HTML}{EF7F29}
\definecolor{lightorange}{HTML}{FCE8DD}

\title{Pairwise Similarity Regularization for Semi-supervised Graph Medical Image Segmentation}

\author{
Jialu Zhou$^1$
\and
Dianxi Shi$^{2, }$\footnote{Corresponding author.}\and
Shaowu Yang$^1$\and
Chunping Qiu$^3$\and
Luoxi Jing$^4$\and
Mengzhu Wang$^{5}$\\
\affiliations
$^1$College of Computer Science and Technology, National University of Defense Technology, China\\
$^2$Academy of Military Sciences, China\\
$^3$Intelligent Game and Decision Lab, China\\
$^4$Peking University, China \\
$^5$Hebei University of Technology, China\\
\emails
\{zhoujialu23, dxshi, shaowu.yang\}@nudt.edu.cn,\\
chunping.qiu@aliyun.com,
jingluoxi@stu.pku.edu.cn,
dreamkily@gmail.com,
}
\begin{document}

\maketitle

\begin{abstract}
    With fully leveraging the value of unlabeled data, semi-supervised medical image segmentation algorithms significantly reduces the limitation of limited labeled data, achieving a significant improvement in accuracy. However, the distributional shift between labeled and unlabeled data weakens the utilization of information from the labeled data. To alleviate the problem, we propose a graph network feature alignment method based on pairwise similarity regularization (PaSR) for semi-supervised medical image segmentation. PaSR aligns the graph structure of images in different domains by maintaining consistency in the pairwise structural similarity of feature graphs between the target domain and the source domain, reducing distribution shift issues in medical images. Meanwhile, further improving the accuracy of pseudo-labels in the teacher network by aligning graph clustering information to enhance the semi-supervised efficiency of the model. The experimental part was verified on three medical image segmentation benchmark datasets, with results showing improvements over advanced methods in various metrics. On the ACDC dataset, it achieved an average improvement of more than 10.66\%.
\end{abstract}

\section{Introduction}

Computer-aided diagnosis (CAD) utilizes image processing, pattern recognition, and data analysis technologies to provide precise and efficient assistance in the doctors' disease diagnosis process. Medical images such as computed tomography (CT) or magnetic resonance imaging (MRI) serve as important references in clinical diagnosis. Precise identification of structures within these images has high value in clinical practice ~\cite{xu2024advances}. However, it’s a great challenging to obtain large-scale, accurately labeled medical datasets for training models. Because manually drawing contours requires professional personnel to go through complex procedures, consuming a lot of resources, and is prone to subjective judgment. In this process, there is a high possibility to lose details leading to a decrease in the accuracy and precision of the segmentation result ~\cite{bai2023bidirectional}. To reduce the dependence on labeled data, semi-supervised medical image segmentation (SSMIS) has emerged as an effective solution. SSMIS enables segmentation models to learn from a small amount of labeled samples and then train with unlabeled samples through methods such as pseudo-label generation, consistency learning, and adversarial learning, thereby fully utilizing the knowledge information in various types of data and improving the accuracy of segmentation.

Although the distribution of labeled and unlabeled data in medical images is consistent in theory, due to the extremely limited nature of labeled data, it is challenging to accurately infer the true distribution of the data. This usually leads to the problem of distribution shift between unlabeled data and labeled data with large data volume differences ~\cite{wang2019semi}. The researchers use the semi-supervised learning method to integrate unlabeled data resources to assist the model in mastering the inherent structure and probability distribution of the data. For example, Inf-Net~\cite{fan2020inf} aggregated advanced features through parallel partial attention, and then used implicit reverse attention and explicit edge attention to model boundaries and enhance representations. BCP \cite{bai2023bidirectional} proposed a bidirectional copy-paste method, which enhances the consistency of data by pasting the foreground taken from the labeled or unlabeled data onto the background of another type of data for data mixing. GraphCL \cite{wang2024graphcl} jointly models graph data structures and introduces structural information into the image segmentation process, achieving better segmentation results. Previous studies have largely addressed the issue of data quantity differences from the adaptations of domain adaptation, pseudo-label generation, or data augmentation, but less so from the alignment of graph structural information to mitigate the impact of distribution shift on segmentation accuracy.

Feature alignment provides a solution to the distribution shift problem between datasets by aligning the features of different domains to a common feature space, thereby improving the model's generalization ability in the target domain. It is also applied in medical image segmentation to address the domain differences between labeled and unlabeled data. For example, Yu et al. \cite{yu2023source} using pre-trained pixel-level classifier weights as the source prototype, minimize the expected cost to align the target features with category prototypes, thereby significantly improving the performance of medical image segmentation in the target domain. In medical image segmentation, the parts of images in the same category usually represent the same tissue or organ in the human body, which are highly similar in physiological composition and metabolic activity. This similarity leads to a consistent response to device signals during imaging, resulting in higher similarity of features for this category in the image. Therefore, feature similarity information is of great significance in medical image segmentation tasks. But existing research has paid little attention to the role of image feature similarity in aligning unbalanced data distribution.

To solve the aforementioned issues, we propose a graph network alignment method based on pairwise similarity regularization (PaSR). By measuring the similarity within and between domains, the alignment model alleviates the domain distribution bias, while utilizing approximate graph clustering information to guide the teacher network to generate more accurate pseudo-labels, achieving the joint optimization of the teacher-student network. Specifically, we take a voxel feature from an image as the node of the graph structure and pairwise similarity as the weight of the edges. By training the model to measure the similarity of different domain features uniformly, align the graph structure of samples from different domains. Then, through Graph Convolutional Network (GCN), we aggregate the global and local information of the image to generate more generalized features. At the same time, minimizing the pairwise similarity regularization based on intra-domain features can also transfer information from aligned data to the pseudo-label generation process during training, improving the quality of pseudo-labels. Thus the model can take into account the comprehensive performance of the teacher network. The core contributions of this paper are as follows:
\begin{itemize}
    \item We propose a graph network alignment method (PaSR) for SSMIS. By constraining the similarity between feature graph structures, the consistency of output features and the quality of pseudo labels are positively guided. 

    \item We conducted various experiments on several common medical image segmentation benchmarks and achieved satisfactory performance.
\end{itemize}

\section{Related Work}

\subsection{Medical Image Segmentation}
Originating from the significant importance of medical image segmentation in clinical medicine, a large number of excellent works have emerged in this field. There are methods such as federated learning \cite{cai2023fedce}, diffusion models \cite{rahman2023ambiguous}, or defining new directional connectivity \cite{yang2023directional} to improve segmentation accuracy.

\subsection{Semi-supervised Medical Image Segmentation}

Semi-supervised learning aims to improve the effectiveness of supervised learning by integrating rich unlabeled data resources \cite{han2024deep}, providing an efficient solution strategy for medical image segmentation tasks with limited labeled samples. Existing optimization methods can be roughly divided into contrastive learning methods \cite{basak2023pseudo,zhang2023multi}, pseudo-labeling methods \cite{liu2023multi,thompson2022pseudo}, and consistency methods \cite{bai2023bidirectional,lei2022semi} among various approaches. For example, ComWin \cite{wu2023compete} trains multiple models or multiple branches of models to generate pseudo-labels, using a competitive mechanism to motivate the improvement of the quality of pseudo-labels. UMCT \cite{xia2020uncertainty} proposes a multi-view collaborative training method, applying collaborative training by enforcing multi-view consistency on unlabeled data.

\subsection{Graph Feature Alignment}

Feature alignment aims to reduce the distribution differences between source and target domains by adjusting the data feature representations, thereby improving the model's performance in the target domain. BSAFusion\cite{li2024bsafusion} integrates alignment and fusion tasks through a shared feature encoder and enhances the accuracy of feature alignment with a bidirectional stepwise feature alignment strategy. Fapn\cite{huang2021fapn} learns pixel transformation offsets to align upsampled feature maps with reference feature maps. Graph features alignment focuses on the graph structures between features, enabling the alignment of node features while considering the structural relationships between nodes, thus achieving knowledge transfer across graphs. GRASP\cite{hermanns2023grasp} aligns nodes using the functional correspondence derived from the eigenvectors of the Laplacian matrix and achieved good results under high noise levels. SLOTAlign\cite{tang2023robust} converts graph alignment into an optimal transport problem between two intra-graph matrices and combines multi-view structural learning to align graph features without cross-graph comparisons. Feature alignment methods effectively addresses domain shifts in practical applications. By obtaining more consistent feature representations, models can better achieve domain transfer of knowledge, enhancing their performance across various domains in different application fields\cite{dan2024hogda,chen2020unsupervised,luo2023adversarial,duan2023efecl}.
 
\begin{figure*}      
	\centering
    \includegraphics[scale=0.38]{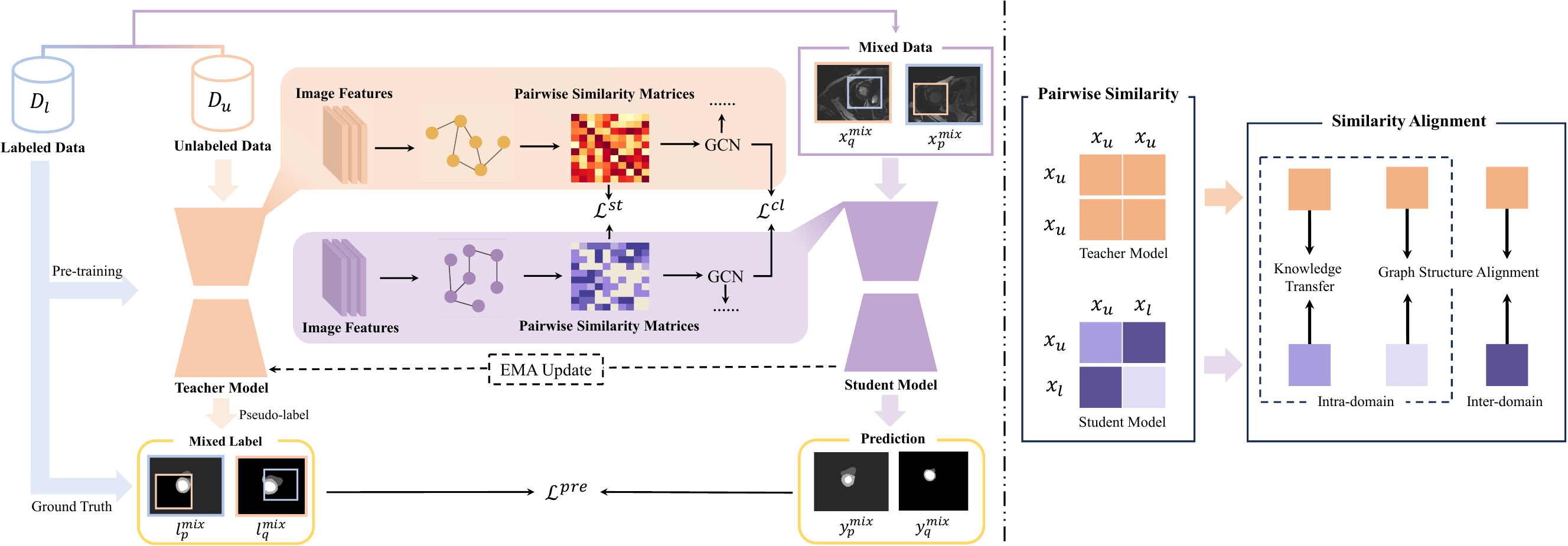}  %scale缩放比例，Fig.jpg文件名
	\caption{The left figure shows the overall framework of the model. A graph structure is constructed with each voxel feature of the medical images that have been copied and pasted bidirectionally as nodes. The similarity between node features is calculated as the pairwise similarity matrix of the graph, which serves as the weight of the edges. Then, the aligned pairwise similarity matrix is used as the adjacency matrix, inputted into the GCN for information aggregation. Finally the graph clustering information is used to assist in the segmentation task. The right figure shows the process of pairwise similarity regularization alignment. By reducing the pairwise similarity distance between the teacher-student networks, it aligns the intra-domain and inter-domain similarities of different domain features. Thus guiding the network to generate a more consistent graph structure and transfering the aligned image knowledge to pseudo-label generation.}   % 图片名称
	\label{framework}
\end{figure*}

\section{Method}

\subsection{Problem Definition}

In the medical image segmentation tasks, we typically employ the following formal definitions to describe the training dataset and objectives. Let the training dataset $D$ consist of two parts: the labeled dataset $D_l$ and the unlabeled dataset $D_u$. The labeled dataset $D_l$ contains $N_l$ samples along with their corresponding label pairs, denoted as $D_l = \{(X^l, Y^l)\} $. The unlabeled dataset $D_u$ comprises $N_u$ samples without labels, denoted as $D_u = \{(X^u)\} $. For each sample, the three-dimensional volume is defined as $X^l, X^u \in \mathbb{R}^{C \times D \times W \times H}$, where $C$ represents the number of channels, and $D, H$, and $W$ represent depth, height, and width, respectively. The sample labels are defined as
$Y^l \in \mathbb{R}^{C \times D\times W \times H}$.

The objective is to predict the label map for voxels, which is formulated as $Y^p \in \{0, 1, \ldots, M-1\}^{C \times D \times W \times H}$, where $M$ is the total number of classes. The task aims to minimize the discrepancy between the predicted label map $Y^p$ and the labels from dataset $Y^l$ thereby achieving as accurate a segmentation effect as possible.

PaSR improves the encoder of the teacher-student network. First, we use CNN to obtain the feature values of the network input images, calculate the pairwise similarity matrix of the features and construct a weighted undirected graph of voxels. Pairwise similarity regularization can align the inter-domain and intra-domain similarities in the pairwise similarity matrices of the unlabelled and mixed images. By minimizing the differences in the model's measurement of feature similarity across different domains, alleviate the distributional shift between unlabeled and labeled data, improve the quality of pseudo-labels generated by the teacher network, and provide more generalized graph structural features for the GCN network. The framework of PaSR is showed in Figure \ref{framework}.

\subsection{Bidirectional Copy-Paste Framework}
The Bidirectional Copy-Paste Framework employs knowledge distillation for semi-supervised training of the model. This framework includes a teacher network $ \mathcal{N}_T(x_i^u, x_j^u; \Theta_t) $ and a student network $ \mathcal{N}_S(x^{mix}_p, x^{mix}_q; \Theta_s) $, where $\Theta_t $ and $\Theta_s $ are the model parameters. $x^{mix}_p,x^{mix}_q$ are the mixed samples obtained through bidirectional copy-paste. Specifically, this involves randomly cropping a foreground from both labeled and unlabeled images and pasting it onto the background of another part of the dataset. The process is as follows: first, two images are randomly selected from the labeled data and unlabeled data, respectively, $x_a^l, x_b^l \in X^l, a\neq b$ and $x_s^u, x_t^u \in X^u, s \neq t$. Then a region is randomly selected from $x_a^l$ and $x_t^u$ as the foreground and pasted onto the background of $x_s^u$ and $x_b^l $ to obtain $x^{mix}_q$ and $x^{mix}_p$. The determination of the cropping region can be achieved using a mask $\mathcal{S} \in \{0, 1\}^{C \times D \times W \times H}$, where the value at each voxel's corresponding position indicates whether the voxel is retained in the mixed image. The regions masked are cropped, and the remaining image is retained. The process is as follows:

\begin{equation}  
x^{mix}_p = x_b^l \cdot \mathcal{S} +x_t^u \cdot (1 - \mathcal{S}) 
\end{equation}
\begin{equation}  
x^{mix}_q =x_s^u \cdot \mathcal{S} + x_a^l \cdot (1 - \mathcal{S})
\end{equation}

The training of teacher-student networks can be divided into two phases: pre-training and self-training. When using mixed images for model training, taking sample labels from the dataset as supervision labels is inappropriate. Instead, a composite label is needed that amalgamates both labeled and unlabeled images. Therefore, we need to generate the pseudo-labels for unlabeled images to create mixed labels. This task is delegated to the teacher network. In the pre-training phase, the teacher network is trained on labeled images to produce pseudo-labels as close to ground truth as possible. Following this, during the self-training phase, the pre-trained teacher network generates pseudo-labels for unlabeled data:
\begin{equation}  
L^p = \mathcal{N}_T (X^u; \Theta_t)
\end{equation}

After acquiring pseudo-labels for unlabeled images, a copy-paste method similar to image mixing can be utilized to blend these pseudo-labels with the labels from dataset to produce mixed labels:
\begin{equation}  
l^{mix}_ p= y_b^l \cdot \mathcal{S} + l_t^p \cdot (1 - \mathcal{S})
\end{equation}
\begin{equation}
l^{mix}_q = l_s ^p \cdot \mathcal{S} + y_a^l  \cdot (1 - \mathcal{S})
\end{equation}
where $y_a^l, y_b^l \in Y^l$ are the labels for labeled images $x_a^l, x_b^l$ and $l_t^p, l_s ^p \in L^p$ are the pseudo-labels for unlabeled images $x_t^u,x_s^u$. During the self-training phase, the student network is trained on mixed images to generate predicted results:
\begin{equation}  
Y^{mix}= \mathcal{N}_S(X^{mix}; \Theta_s)
\end{equation}
where $ x^{mix}_p, x^{mix}_q \in X^{mix}$ are the collection of mixed images and $ Y^{mix}$ is the corresponding predictions. Finally, construct the loss function for $Y^{mix}$ and the mixed labels $L^{mix}$, then update the model parameters, see details in section \hyperlink{section1}{3.4}.

\subsection{Pairwise Similarity Alignment}

Although theoretically the data distribution of labeled and unlabeled medical images should be consistent, the scarcity of labeled data often leads to distribution shifts between the two.
In order to fully utilize the value of similarity information in feature alignment \cite{wang2020pairwise}, we propose a graph network alignment method based on pairwise similarity. By reducing the pairwise similarity regularization between mixed samples and unlabeled samples, we achieve joint optimization of the teacher-student network. PaSR aligns the model to evaluate cross-domain and intra-domain data similarity, harmonizes feature map structures, and mitigates distribution bias. Meanwhile, PaSR introduces knowledge from aligned data into the teacher model learning process through approximate clustering information alignment, making it generate more accurate pseudo-labels and improving the overall semi-supervised efficiency of the model.

PaSR only modify the encoder of the backbone. We first need to extract features from the input image $X_{in}$ which is accomplished through pooling layers in the CNN:
\begin{equation}  
\mathcal{F}= P_{CNN}(X_{in}; \Theta_C)
\end{equation}
where $\Theta_C$ is the parameter of CNN and extracted features $\mathcal{F} \in \mathbb{R}^{B \times V}$. $B$ denotes batch size and $V$ denotes the dimension of the voxel in a 3D image. In medical images, voxels of the same class often have similar properties, which translates to higher feature similarity in the feature space. Therefore, we construct a pairwise similarity matrix by calculating the feature similarity within the network:
\begin{equation}  
\mathcal{A}_u= \mathcal{F}_u \cdot \mathcal{F}_u^\intercal-\frac{MAX(\mathcal{F}_u \cdot \mathcal{F}_u^\intercal)}{\mu} 
\end{equation}
\begin{equation}  
\mathcal{A}_m = \mathcal{F}_m \cdot \mathcal{F}_m^\intercal-\frac{MAX(\mathcal{F}_m \cdot \mathcal{F}_m^\intercal)}{\mu}
\end{equation}
where $\mathcal{A} \in \mathbb{R}^{B \times N \times N}$, $N$ represents the number of nodes. $\mathcal{F}_u, \mathcal{F}_m \in \mathcal{F}$ are the image features of the unlabeled image and the mixed image. $\mu$ scales the similarity matrix to accommodate subsequent clustering tasks. 

Based on $\mathcal{A}$, we construct a weighted undirected graph structure with voxel features, where voxel features are taken as nodes and the similarity between nodes as the weight of the edges. Because nodes with higher similarity are more likely to be clustered together in subsequent clustering tasks, we can approximate the clustering results using $\mathcal{A}$ which can also preserve the network structure of nodes across different clusters. PaSR calculate the average distance between the teacher and student networks' $\mathcal{A}$ as pairwise similarity regularization:
\begin{equation}  
d(\mathcal{A}_u,\mathcal{A}_m)=\frac{1}{N^2}\sum_{i=1}^{N}\sum_{j=1}^{N} |\mathcal{A}_{u _{ij}} -  \mathcal{A}_{m _{ij}}|
\end{equation}
where $ \mathcal{A}_{u _{ij}}, \mathcal{A}_{m _{ij}}$ represents the similarity between the $i$th voxel feature and the $j$th voxel feature. 

Mixed images are composed of labeled and unlabeled data, which means $\mathcal{A}_m$ can reflect both inter-domain and intra-domain feature similarity. In the matrix $\mathcal{A}_{m_{ij}}$, when features $i$ and $j$ are both from unlabeled data, the student network transfers the knowledge learned from aligned data to the teacher network's segmentation process through approximate clustering information, generating higher quality pseudo labels. When features $i$ and $j$ come from different data subsets, $\mathcal{A}_m$ and $\mathcal{A}_u$ represent inter-domain and intra-domain feature similarities, respectively. By shortening the average distance between them, the model can be guided to measure the similarity between cross-domain features to approach the intra-domain features. Distribution alignment is achieved by aligning the graph structures from inter-domain and intra-domain data. When features $i$ and $j$ are both from labeled data, aligning the intra-domain similarities between domains can reduce the distribution shifts. Based on this, PaSR jointly optimizes the teacher-student network through pairwise similarity regularization, improving the quality of the teacher network's pseudo labels while alleviating distribution shifts between datasets:
\begin{equation}  
\Theta_t \leftarrow \Theta_t - r \nabla_{\Theta_t} d(\mathcal{A}_u,\mathcal{A}_m)
\end{equation}
\begin{equation}  
\Theta_s \leftarrow \Theta_s - r \nabla_{\Theta_s} d(\mathcal{A}_u,\mathcal{A}_m)
\end{equation}
where $r$ is the learning rate. To fully utilize the spatial structure of features, we take the pairwise similarity matrix as the adjacency matrix and use GCN to extract the features of graph structure $\mathcal{G}$. Taking $\mathcal{G}$ and $\mathcal{A}$ as the input of GCN, $\mathcal{G}$ is mapped to a more refined feature matrix $\hat{\mathcal {G}}$ with the relationship structure in $\mathcal{A}$:
\begin{equation}  
\hat{\mathcal{G}}=GCN(\mathcal{A} , \mathcal{G};\Theta_G)
\end{equation}
where $\Theta_G $ is the parameter of GCN. $\hat{\mathcal{G}}$ aggregates the characteristics of neighboring nodes, which can better capture global and local information. In addition, we use a Multi-Layer Perceptron(MLP) based on soft-max to implement clustering of graph nodes represented by  $\hat{\mathcal{G}}$ to further supervise the segmentation of graph nodes:
\begin{equation}  
\mathcal{C}=MLP(\hat{\mathcal{G}};\Theta_M)
\end{equation}
where $\Theta_M $ is a parameter of MLP. $\mathcal{C}_{ij}$ is the value in matrix $\mathcal{C}$ which is the probability that node $i$ is assigned to cluster $j$.
\hypertarget{section1}{}
\subsection{Loss Function and Model Update}
%\subsection{Loss Function and Model Update}
\textbf{Model prediction loss.} The prediction loss is articulated using the weighted fusion of Dice loss and Cross-Entropy loss for predictions of mixed images  $ y^{mix}_p, y^{mix}_q \in Y^{mix}$ and mixed labels $ l^{mix}_ p, l^{mix}_q $. Given that the model's input is a mixture of labeled and unlabeled images, and considering the varying reliability of label values between the two, it’s unsuitable to assign equal contribution to both parts in the loss function. Therefore, $\gamma$ is used to adjust the weight of different parts of the loss within the overall loss.
\begin{align}
 \mathcal{L}^{pre}_p = \mathcal{L}_{com} &\left( l^{mix}_ p, y^{mix}_p \right) \cdot \mathcal{S} + \nonumber \\& \gamma \mathcal{L}_{com} \left( l^{mix}_ p, y^{mix}_p \right) \cdot (1 - \mathcal{S}) \quad
\end{align}
\begin{align}
    \mathcal{L}^{pre}_q = \mathcal{L}_{com} ( l^{mix}_q, &y^{mix}_q ) \cdot (1 - \mathcal{S}) + \nonumber \\& \gamma \mathcal{L}_{com} \left(l^{mix}_q, y^{mix}_q \right) \cdot \mathcal{S } \quad
\end{align}
\begin{align}
    \mathcal{L}^{pre} = \mathcal{L}^{pre}_p+\mathcal{L}^{pre}_q
\end{align}

{
\setlength{\parindent}{0cm}
\textbf{Pairwise similarity regularization loss.} This loss is calculated based on pairwise similarity between two networks. We use the average distance to measure the overall similarity of the image graph structure. Therefore, the pairwise similarity loss is computed as follows:
\begin{equation}  
\mathcal{L}^{st} =d(\mathcal{A}_u,\mathcal{A}_m)=\frac{1}{N^2}\sum_{i=1}^{n}\sum_{j=1}^{n}|A_{u _{ij}} - A_{m _{ij}}|
\end{equation}
}

{
\setlength{\parindent}{0cm}
\textbf{Graph clustering loss.} The last loss in the model is to calculate the clustering loss for the clustering results $\mathcal{C}$. We use correlation clustering loss to achieve the clustering goal by increasing the probability of nodes with high feature similarity being assigned to the same cluster and decreasing the probability of connections between nodes with low similarity:
\begin{equation}  
\mathcal{L}^{cl} =-Tr(\mathcal{A}\mathcal{C}\mathcal{C}^\intercal)
\end{equation}
}
In summary, the overall loss of the model consists of the following three parts:
\begin{equation}  
\mathcal{L} =\mathcal{L}^{pre}+\alpha\mathcal{L}^{st}+\beta\mathcal{L}^{cl}
\end{equation}
where $\alpha$ and $\beta$ are used to control the contribution of graph domain alignment and graph clustering to model training. 

The parameter updates of the entire network framework utilize the Exponential Moving Average (EMA) mechanism. That is, the student network updates its parameters through backpropagation, while the teacher network's parameters are updated via the EMA method instead of direct gradient descent, ensuring a stable parameter evolution:
\begin{align}
\Theta_{t}^T = \lambda \cdot \Theta_{t}^{T-1} + (1 - \lambda) \cdot \Theta_{s}^T
\end{align}
where $\lambda$ is a smoothing coefficient, $\Theta_{t}^T$ represents the parameters of the teacher network at the $T$-th iteration.

\section{Experiment}

\subsection{Evaluation Metrics}
We utilize Dice Score, Jaccard Score, 95\% Hausdorff Distance (95HD), and Average Surface Distance (ASD) to assess our experimental outcomes. In particular, Dice Score and Jaccard Score serve to gauge the resemblance between predictions and actual results, where values nearing 1 denote enhanced similarity. Meanwhile, 95HD and ASD correspond to the maximal and mean surface distances between the prediction and the ground truth. For these metrics, lower readings equate to improved segmentation precision.

\subsection{Datasets and Implementation Details}
In the absence of specific instructions, the default settings $\gamma=0.5$, $\beta=0.01$, $\mu=2$ with fixed random seeds are utilized in all experiments. All experiments use an NVIDIA 4090 GPU.

{
\setlength{\parindent}{0cm}
\textbf{PROMISE12.} The PROMISE12 dataset, a medical image collection for prostate MRI segmentation \cite{litjens2014evaluation}, comprises T2-weighted MRI images from 50 patients. In this study, all datasets are partitioned into training, validation, and test sets at a 7 : 2 : 1 ratio. The data is three-dimensional. A batch size of 12 is used, with 10,000 pre-training and 30,000 self-training iterations. The model was trained and tested on 20\% labeled data, with $\alpha$ set at 0.05.
}

{
\setlength{\parindent}{0cm}
\textbf{ACDC.} Automatic Cardiac Diagnosis Challenge(ACDC) is a challenge dataset specifically used for cardiac MRI image analysis and automatic diagnosis \cite{bernard2018deep}. The dataset includes 150 cases, divided into 5 subcategories of normal and abnormal, each containing 20 cases for training. The data dimension is 3D.  The batch size is 12, with 10000 and 30000 iterations for pre-training and self-training, respectively. While training with 5\% and 10\% labeled datasets, $\alpha$ is set to 0.05 and 0.01. For PROMISE12 and ACDC, we refer to SS-Net \cite{wu2022exploring}, segmenting samples in 2D (piece by piece) and using 2D U-Net \cite{ronneberger2015u}  as the backbone network. The input patch size is 256 × 256 and the size of mask $\mathcal{S}$ is 170 × 170. And using SGD as the optimizer with learning rate 0.01 for init.
}

{
\setlength{\parindent}{0cm}
\textbf{Pancreas-NIH.} Pancreas-NIH dataset is provided by the National Institutes of Health (NIH) \cite{roth2015deeporgan}, containing 82 cases of abdominal enhanced 3D CT scan data. We refer to CoraNet \cite{shi2021inconsistency}, using V-Net\cite{milletari2016v} as the backbone network, with input patch size of 96 × 96 × 96 and mask $\mathcal{S}$ size of 64 × 64 × 64. Batch size and training epochs for pre-training and self-training are set to 2, 100, and 200, respectively. For 10\% and 20\% labeled datasets, $\alpha$ is taken as 0.05. Using Adam \cite{diederik2014adam} as the optimizer with learning rate 0.001 for init.
}

\begin{table}[]
\centering
%\tiny
\scriptsize
\begin{tabular}{c|cc|cccc}
\hline
                         & \multicolumn{2}{c|}{Scans used}                        & \multicolumn{4}{c}{Metrics}                                                                                                  \\ \cline{2-7} 
\multirow{-2}{*}{Method} & Labeled                   & Unlabeled                  & DICE↑                         & Jaccard↑                      & 95HD↓                         & ASD↓                         \\ \hline
U-Net                    &                           &                            & 62.28                         & 50.42                         & 16.55                         & 3.56                         \\
DTC                      &                           &                            & 72.03                         & 58.32                         & 11.48                         & \cellcolor[HTML]{D0D8EB}2.65 \\
MC-Net                   &                           &                            & 71.77                         & 59.07                         & 10.76                         & 2.85                         \\
SS-Net                   &                           &                            & 71.56                         & 59.35                         & 14.38                         & 3.03                         \\
BCP                      &                           &                            & 81.71                         & 69.28                         & 11.88                         & 7.45                         \\
GraphCL                  & \multirow{-6}{*}{7(20\%)} & \multirow{-6}{*}{28(80\%)} & \cellcolor[HTML]{D0D8EB}82.01 & \cellcolor[HTML]{D0D8EB}70.60 & \cellcolor[HTML]{D0D8EB}10.47 & 6.74                         \\ \hline
\textbf{PaSR}            &                           &                            & \cellcolor[HTML]{7195C5}82.90 & \cellcolor[HTML]{7195C5}71.28 & \cellcolor[HTML]{7195C5}4.87  & \cellcolor[HTML]{7195C5}2.46 \\ \hline
\end{tabular}
\caption{Comparison experiment results on the PROMISE12. The best results are indicated by dark, and the suboptimal results are light.}
\label{PROMISE12}
\end{table}

\subsection{Comparison with Semi-supervised Methods}

Except for the corresponding backbone network U-Net and V-net, the baselines in ACDC and Pancreas-NIH are UA-MT\cite{yu2019uncertainty}, SASSNet\cite{li2020shape}, DTC\cite{luo2021semi}, SS-Net\cite{wu2022exploring}, BCP\cite{bai2023bidirectional} and GraphCL\cite{wang2024graphcl}. On the PROMISE12, we compared our method with DTC\cite{luo2021semi}, MC-Net\cite{wu2021semi}, SS-Net\cite{wu2022exploring}, BCP\cite{bai2023bidirectional} and GraphCL\cite{wang2024graphcl}. In the experiments, U-Net and V-Net are trained only with the labeled samples.
{
\setlength{\parindent}{0cm}

\textbf{PROMISE12.} Table\ref{PROMISE12} shows the segmentation results on a 20\% labeled dataset. Our method outperforms the other method on all metrics. DICE and Jaccard improve by 1.08\% and 0.96\%, respectively. There is a significant improvement in the 95HD and ASD metrics, with increases of 53.48\% and 7.17\%. All metrics improve on average by 15.68\%. The improvement in model performance is due to the graph feature alignment guiding the model to learn more generalized feature representations, better capturing the graph structure in the image and inputting it into the GCN. Additionally, the inclusion of the pairwise similarity regularizer makes the model not only focus on the loss of labeled data but also on the quality of pseudo-label generation, which helps to enhance the model's stability and robustness.
}

\begin{table}[]
\centering
%\tiny
\scriptsize
\begin{tabular}{c|cc|cccc}
\hline
                         & \multicolumn{2}{c|}{Scans used}                        & \multicolumn{4}{c}{Metrics}                                                                                                 \\ \cline{2-7} 
\multirow{-2}{*}{Method} & Labeled                   & Unlabeled                  & DICE↑                         & Jaccard↑                      & 95HD↓                        & ASD↓                         \\ \hline
U-Net                    &                           &                            & 47.64                         & 36.98                         & 31.21                        & 12.64                        \\
UA-MT                    &                           &                            & 46.04                         & 35.97                         & 20.08                        & 7.75                         \\
SASSNet                  &                           &                            & 57.77                         & 46.14                         & 20.05                        & 6.06                         \\
DTC                      &                           &                            & 56.90                         & 45.67                         & 23.36                        & 7.39                         \\
SS-Net                   &                           &                            & 65.83                         & 55.38                         & 6.67                         & 2.28                         \\
BCP                      &                           &                            & 87.96                         & 79.18                         & \cellcolor[HTML]{D0D8EB}4.14 & \cellcolor[HTML]{D0D8EB}1.14 \\
GraphCL                  & \multirow{-7}{*}{3(5\%)}  & \multirow{-7}{*}{67(95\%)} & \cellcolor[HTML]{D0D8EB}88.18 & \cellcolor[HTML]{D0D8EB}79.61 & 4.64                         & 1.34                         \\ \hline
\textbf{PaSR}            &                           &                            & \cellcolor[HTML]{7195C5}88.91 & \cellcolor[HTML]{7195C5}80.60 & \cellcolor[HTML]{7195C5}3.14 & \cellcolor[HTML]{7195C5}0.80 \\ \hline
U-Net                    &                           &                            & 79.38                         & 68.09                         & 9.41                         & 2.76                         \\
UA-MT                    &                           &                            & 81.65                         & 70.64                         & 6.88                         & 2.02                         \\
SASSNet                  &                           &                            & 84.50                         & 74.34                         & 5.42                         & 1.86                         \\
DTC                      &                           &                            & 84.29                         & 73.92                         & 12.81                        & 4.01                         \\
SS-Net                   &                           &                            & 86.78                         & 77.67                         & 6.07                         & 1.40                         \\
BCP                      &                           &                            & \cellcolor[HTML]{D0D8EB}89.03 & \cellcolor[HTML]{D0D8EB}80.84 & 5.12                         & 1.19                         \\
GraphCL                  & \multirow{-7}{*}{7(10\%)} & \multirow{-7}{*}{63(90\%)} & 88.06                         & 79.50                         & \cellcolor[HTML]{D0D8EB}3.63 & \cellcolor[HTML]{D0D8EB}1.13 \\ \hline
\textbf{PaSR}            &                           &                            & \cellcolor[HTML]{7195C5}89.43 & \cellcolor[HTML]{7195C5}81.47 & \cellcolor[HTML]{7195C5}3.09 & \cellcolor[HTML]{7195C5}0.83 \\ \hline
\end{tabular}
\caption{Comparison experiment results on the ACDC. he best results are indicated by dark, and the suboptimal results are light.}
\label{ACDC}
\end{table}
{
\setlength{\parindent}{0cm}
\textbf{ACDC.} Table \ref{ACDC} shows the segmentation results of the four-class on the ACDC dataset with labeled ratios of 5\% and 10\%. Our method outperforms the other methods on all metrics. On the 5\% labeled dataset, Dice and Jaccard improve by 0.83\% and 1.24\% respectively compared to the second-best method. Our method showes significant improvements in the 95HD and ASD metrics, with increases of 24.15\% and 29.82\%. On average, all metrics improve by 14.01\%. On the 10\% labeled dataset, all evaluation metrics are better than the second-best model, with an average improvement of 10.66\%, and significant improvements are also observed in the 95HD and ASD. It can be seen that the improvement effect is more pronounced when there is less labeled data. This shows that our method can better obtain effective information from unlabeled data for model training to make up for the scarcity of labeled data.
}

\begin{table}[]
\centering
%\tiny
\scriptsize
\begin{tabular}{c|cc|cccc}
\hline
                         & \multicolumn{2}{c|}{Scans used}                         & \multicolumn{4}{c}{Metrics}                                                                                                 \\ \cline{2-7} 
\multirow{-2}{*}{Method} & Labeled                    & Unlabeled                  & DICE↑                         & Jaccard↑                      & 95HD↓                        & ASD↓                         \\ \hline
V-Net                    &                            &                            & 55.60                         & 41.74                         & 45.33                        & 18.63                        \\
UA-MT                    &                            &                            & 66.34                         & 53.21                         & 17.21                        & 4.57                         \\
SASSNet                  &                            &                            & 68.78                         & 53.86                         & 19.02                        & 6.26                         \\
DTC                      &                            &                            & 69.21                         & 54.06                         & 17.21                        & 5.95                         \\
SS-Net                   &                            &                            & 71.76                         & 57.05                         & 17.56                        & 5.77                         \\
BCP                      & \multirow{-6}{*}{6(10\%)}  & \multirow{-6}{*}{56(90\%)} & 79.28                         & 66.92                         & 9.14                         & \cellcolor[HTML]{D0D8EB}3.05 \\
GraphCL                  &                            &                            & \cellcolor[HTML]{D0D8EB}81.43 & \cellcolor[HTML]{D0D8EB}68.17 & \cellcolor[HTML]{D0D8EB}8.91 & 3.21                         \\ \hline
\textbf{PaSR}            &                            &                            & \cellcolor[HTML]{7195C5}81.53 & \cellcolor[HTML]{7195C5}69.07 & \cellcolor[HTML]{7195C5}8.80 & \cellcolor[HTML]{7195C5}2.96 \\ \hline
V-Net                    &                            &                            & 72.38                         & 58.26                         & 19.35                        & 5.89                         \\
UA-MT                    &                            &                            & 77.26                         & 63.82                         & 11.90                        & 3.06                         \\
SASSNet                  &                            &                            & 77.66                         & 64.08                         & 10.93                        & 3.05                         \\
DTC                      &                            &                            & 78.27                         & 64.75                         & 8.36                         & 2.25                         \\
SS-Net                   &                            &                            & 78.98                         & 66.32                         & 8.86                         & \cellcolor[HTML]{D0D8EB}2.01 \\
BCP                      & \multirow{-6}{*}{12(20\%)} & \multirow{-6}{*}{50(80\%)} & 81.04                         & 68.73                         & \cellcolor[HTML]{D0D8EB}8.25 & 2.34                         \\
GraphCL                  &                            &                            & \cellcolor[HTML]{D0D8EB}82.47 & \cellcolor[HTML]{D0D8EB}70.46 & 8.72                         & 2.62                         \\ \hline
\textbf{PaSR}            &                            &                            & \cellcolor[HTML]{7195C5}83.54 & \cellcolor[HTML]{7195C5}72.02 & \cellcolor[HTML]{7195C5}4.69 & \cellcolor[HTML]{7195C5}1.73 \\ \hline
\end{tabular}
\caption{Comparison experiment results on the Pancreas-NIH. he best results are indicated by dark, and the suboptimal results are light.}
\label{Pancreas-NIH}
\end{table}
{
\setlength{\parindent}{0cm}
\textbf{Pancreas-NIH.} From the results shown in the Table \ref{Pancreas-NIH}, it can be seen that the our method is superior to the suboptimal method on all metrics. On the 10\% labeled dataset, the metrics are improved by 1.41\% in average. On the 20\% labeled dataset, the metrics improve by 1.29\%, 2.21\%, 43.15\%, and 13.93\%. The performance improvement of our method in the 95HD and ASD indicators is still significant.
}

\subsection{Ablation Studies}
\begin{figure}      
	\centering
    \includegraphics[scale=0.35]{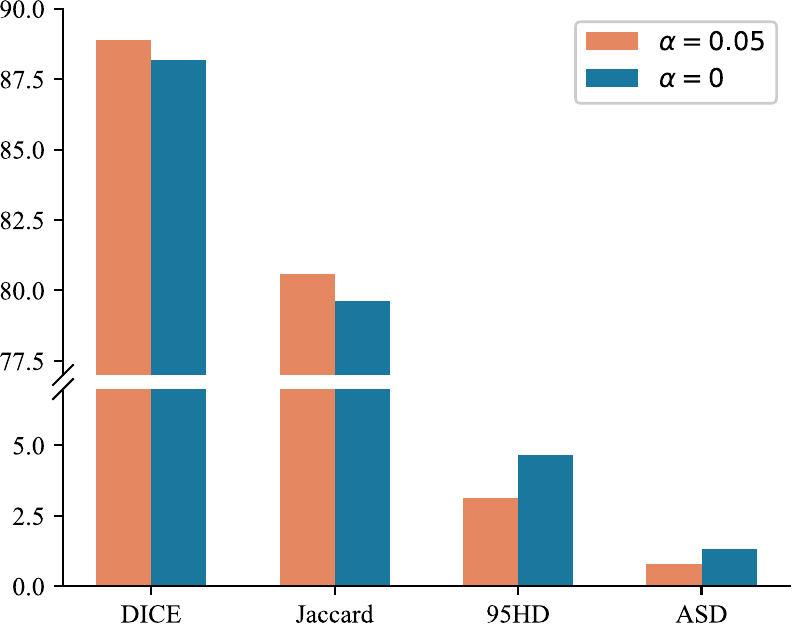 }  %scale缩放比例，Fig.jpg文件名
	\caption{PaSR ($\alpha$=0.05) and model variant ($\alpha$=0) in segmentation performance on the ACDC dataset with 5\% labeled data.}   % 图片名称
    \label{ablation}
\end{figure}

Firstly, to verify the contribution of the graph feature alignment function, we compared the performance of PaSR ($\alpha=0.05$) and a variant without the graph feature alignment function ($\alpha=0$) on the $5\%$ labeled ACDC. As shown in the figure \ref{ablation}, the model with $\alpha=0.05$ outperformed the variant with $\alpha=0$ in all metrics, indicating that the graph feature alignment function can guide the model to achieve higher segmentation accuracy. This proves the importance of graph feature alignment.

Secondly, To further study the relationship between the features used for the calculation of pairwise similarity matrices and the depth of network layers, we separately use the pairwise similarity matrices of each backbone layer for loss calculation. The results in Table \ref{depth} show that the regularization term obtained by using shallow pairwise similarity matrices for calculation can achieve better performance. On the 10\% ACDC dataset, the shallow pairwise similarity matrices are significantly better than the deeper ones, with an average improvement of 31\%. The reason may be that as the depth of the network increases, the network may encounter smoothing problems, weakening the model's ability to distinguish local features, thereby making the adjacency matrix can’t reflect the features of the source domain and target domain well.

\begin{table}[]
\centering
%\tiny
\scriptsize
\begin{tabular}{c|cc|cccc}
\hline
\multirow{2}{*}{Layers} & \multicolumn{2}{c|}{Scans used}                      & \multicolumn{4}{c}{Metrics}                                     \\ \cline{2-7} 
                        & Labeled                  & Unlabeled                 & DICE↑          & Jaccard↑       & 95HD↓         & ASD↓          \\ \hline
1                       & \multirow{5}{*}{3(5\%)}  & \multirow{5}{*}{67(95\%)} & \textbf{88.91} & \textbf{80.60} & \textbf{3.14} & \textbf{0.80} \\
2                       &                          &                           & 81.02          & 69.78          & 6.52          & 1.78          \\
3                       &                          &                           & 80.85          & 70.98          & 4.03          & 0.94          \\
4                       &                          &                           & 39.31          & 30.89          & 63.51         & 35.17         \\
5                       &                          &                           & 34.34          & 23.92          & 43.40         & 19.371        \\ \hline
1                       & \multirow{5}{*}{7(10\%)} & \multirow{5}{*}{63(90\%)} & \textbf{89.43} & \textbf{81.47} & \textbf{3.09} & \textbf{0.83} \\
2                       &                          &                           & 89.43          & 81.47          & 3.25          & 0.90          \\
3                       &                          &                           & 87.44          & 78.29          & 4.28          & 1.45          \\
4                       &                          &                           & 86.34          & 76.78          & 7.19          & 1.94          \\
5                       &                          &                           & 87.35          & 78.13          & 5.44          & 1.41          \\ \hline
\end{tabular}
\caption{Ablation study on ACDC. The influence of feature extraction depth on the results. The best results are indicated by \textbf{bold}.}
\label{depth}
\end{table}

\subsection{Impacts of Hyperparameters}

\begin{figure}[t]
\centering  %图片全局居中
\subfigure[5\% Labeled]{
\label{Sensitivity analysis-3}
\includegraphics[width=4.0cm,height = 2.8cm]{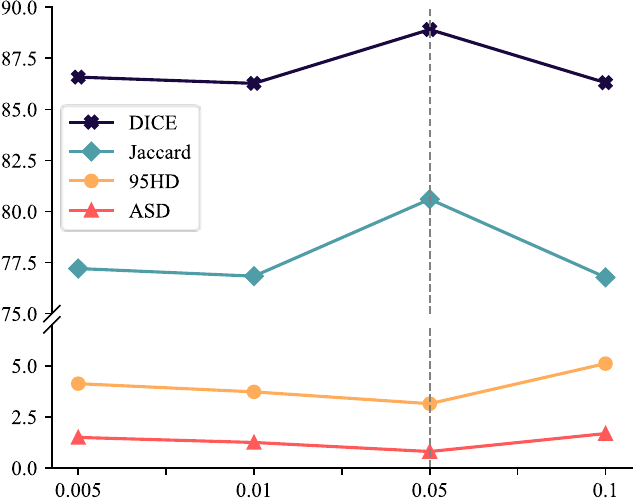}}
\subfigure[10\% Labeled]{
\label{Sensitivity analysis-7}
\includegraphics[width=4.0cm,height = 2.8cm]{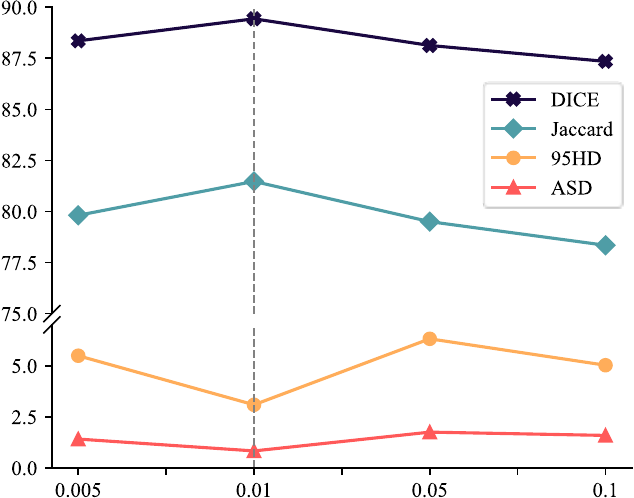}}
\caption{Sensitivity analysis on the ACDC dataset.}
\label{Sensitivity analysis}
\end{figure}

In order to further validate the efficacy of our approach, we conducted a parameter sensitivity analysis on the ACDC dataset to evaluate the impact of the parameter $\alpha$. Figures \ref{Sensitivity analysis-3} and \ref{Sensitivity analysis-7} respectively illustrate the outcomes for varying values of $\alpha$ with 5\% and 10\% labeled data ratios. The parameter $\alpha$ governs the weight of within the model. It is observable that an increase in $\alpha$ can reduce the 95HD and ASD metrics; however, an excessively high value of $\alpha$ can also degrade the overall performance of the model. This may be attributed to the fact that an overemphasis on aligning the source and target domains can lead to overfitting during the feature transformation process, thereby neglecting the critical local features and detailed information in medical image segmentation tasks\cite{pilavci2024graph}. In both datasets with varying labeled proportions, the optimal performance was achieved at $\alpha$ values of 0.05 and 0.01, respectively.

\subsection{Visualization Analysis}

\begin{figure}      
	\centering
    \includegraphics[scale=0.39]{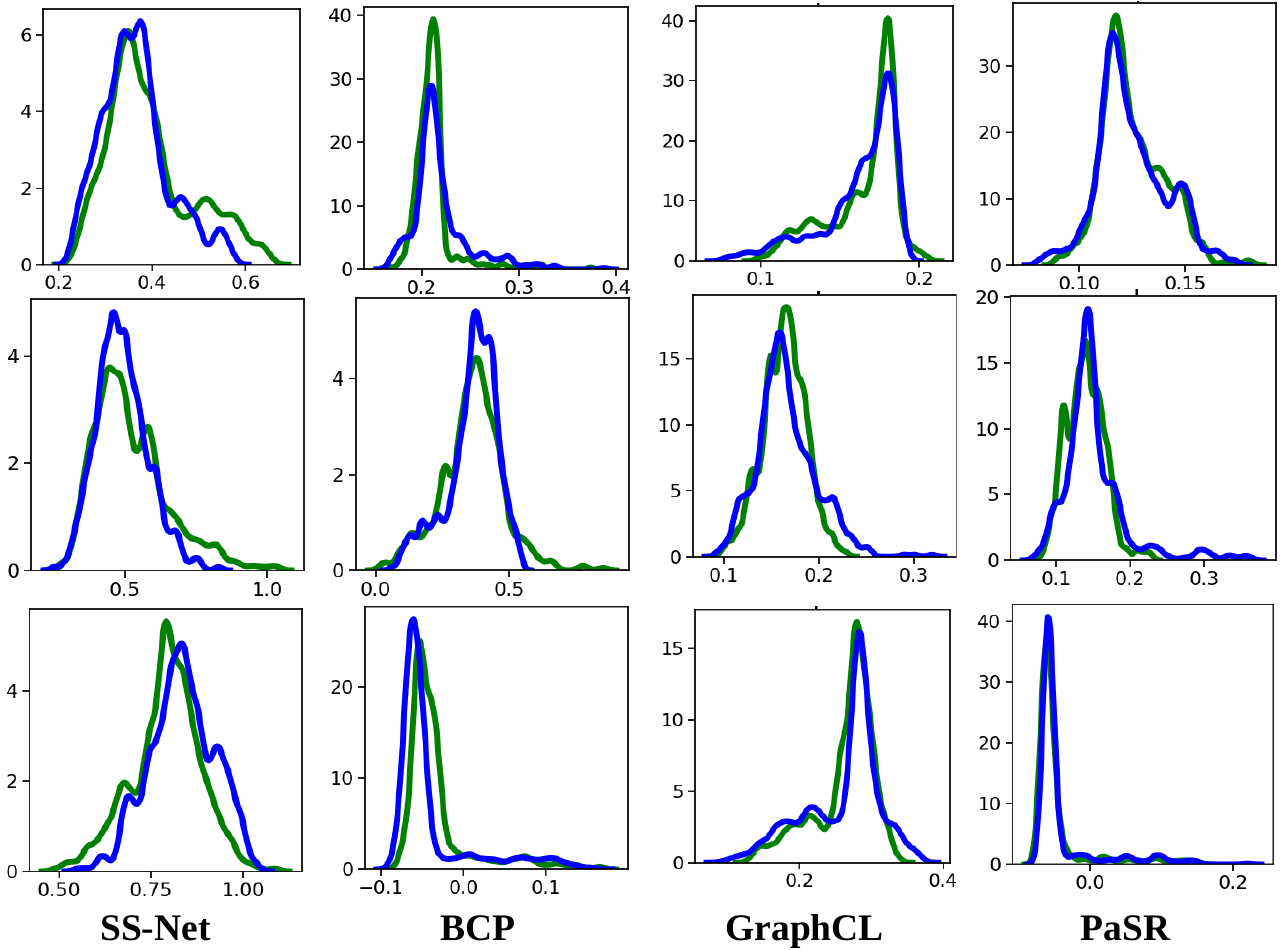}  %scale缩放比例，Fig.jpg文件名
	\caption{Kernel dense estimations of different methods, trained on 10\% labeled ACDC dataset. From top to bottom are the three classes in the ACDC: right ventricle, myocardium and left ventricle. The green and blue line represent the labeled and unlabeled data.}   % 图片名称
	\label{KDE}
\end{figure}

\begin{figure}      
	\centering
    \includegraphics[scale=0.35]{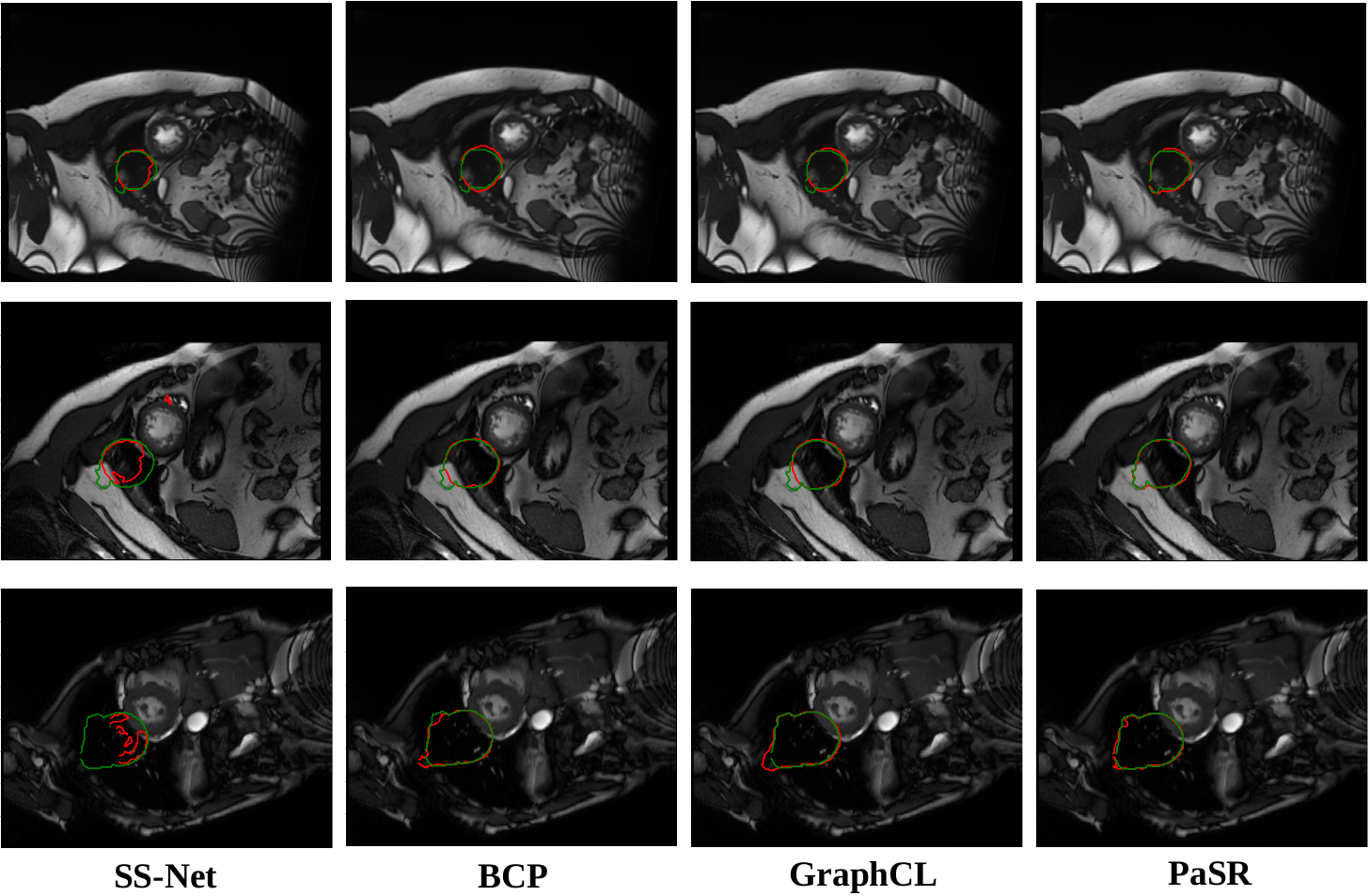 }  %scale缩放比例，Fig.jpg文件名
	\caption{Visualizations of several semi-supervised segmentation methods with 5\% labeled data and ground truth on ACDC dataset. The green and red lines represent the ground truth and prediction results.}   % 图片名称
	\label{Visualization}
\end{figure}

Figures \ref{KDE} and \ref{Visualization} respectively depict the kernel density estimation plots and segmentation result visualizations for diverse training methodologies on the ACDC dataset. Figure \ref{KDE} illustrates that our approach achieves a more effective alignment between labeled and unlabeled data. This suggests that our method, via graph distribution adaptation, more successfully mitigates distribution shift issues across medical datasets. Figure \ref{Visualization} shows that our method's segmentation results are more precise, with a more accurate representation of local details and a closer approximation to the ground truth. These enhancements are credited to the positive guidance of feature acquisition in the target domain, enabled by the transference of knowledge.

\section{Conclusion}
This paper proposes a graph network feature alignment method based on pairwise similarity regularization (PaSR) for SSMIS. We align the similarity measures of the features between domains and within domains so that voxel features from different domains can obtain consistent graph structure. By aligning the feature similarity in unlabeled data domain between networks, the quality of pseudo-label generation is improved by using the knowledge of aligned data. The experiments was validated on the PROMISE12, ACDC, and Pancreas-NIH datasets, showing that our method outperforms advanced methods. On the ACDC dataset, the average improvement exceeded 10.10\%. Future work will explore more complex graph neural network architectures while reducing dependence on hyperparameter tuning, making the model more robust and easy to apply.

\clearpage
\bibliographystyle{named}
\bibliography{ijcai25}

\end{document}